\documentclass[a4paper,twoside]{article}

\usepackage{epsfig}
\usepackage{refstyle}
\usepackage{subfigure}
\usepackage{calc}
\usepackage{amssymb}
\usepackage{amstext}
\usepackage{amsmath}
\usepackage{amsthm}
\usepackage{multicol}
\usepackage{pslatex}
\usepackage{natbib}
\usepackage{SCITEPRESS}
\usepackage{booktabs}
\usepackage{tabularx}
\usepackage{algorithm}
\usepackage{mdwlist}
\usepackage{url}
\usepackage[noend]{algpseudocode}
\usepackage[small]{caption}

\DeclareMathOperator*{\argmax}{arg\,max}
\DeclareMathOperator*{\argmin}{arg\,min}
\newcommand{\W}{\mathcal{W}}

\begin{document}

\title{Learning a Loopy Model For Semantic Segmentation Exactly}



\abstract{
Learning structured models using maximum margin techniques has become an
indispensable tool for computer vision researchers, as many computer vision
applications can be cast naturally as an image labeling problem. Pixel-based or
superpixel-based conditional random fields are particularly popular examples.
Typically, neighborhood graphs, which contain a large number of cycles, are used.
As exact inference in loopy graphs is NP-hard in general, learning these models
without approximations is usually deemed infeasible.
In this work we show that, despite the theoretical hardness, it is possible
to learn loopy models exactly in practical applications.
To this end, we analyze the use of multiple approximate inference techniques
together with cutting plane training of structural SVMs.
We show that our proposed method yields exact solutions with an optimality
guarantees in a computer vision application, for little additional
computational cost.  We also propose a dynamic caching scheme to accelerate
training further, yielding runtimes that are comparable with approximate
methods. We hope that this insight can lead to a reconsideration of the
tractability of loopy models in computer vision.
}

\onecolumn \maketitle \normalsize \vfill

\section{Introduction}
Many classical computer vision applications such as stereo, optical flow, semantic
segmentation and visual grouping can be naturally formulated as image labeling tasks.

Arguably the most popular way to approach such labeling problems is via graphical
models, such as Markov random fields (MRFs) and conditional random fields (CRFs).
MRFs and CRFs provide a principled way of integrating local evidence and
modeling spacial dependencies, which are strong in most image-based tasks.

While in earlier approaches, model parameters were set by hand or using
cross-validation, more recently parameters are often learned using a max-margin
approach.

Most models employ linear energy functions of unary and pairwise interactions,
trained using structural support vector machines (SSVMs). While linear energy
functions lead to learning problems that are convex in the parameters, complex
constraints complicate their optimization. Additionally, inference (or more precisely
loss-augmented prediction) is a crucial part in learning, and can often not be
performed exactly, due to loops in the neighborhood graphs.  
Approximations in the inference then lead to approximate learning.

We look at semantic image segmentation, learning a model of pairwise
interactions on the popular MSRC-21 dataset.
The contribution of this work is threefold:
\begin{itemize}
    \item We analyze the simultaneous use of multiple approximate inference
        methods for learning SSVMs using the cutting plane method, relating
        approximate learning to the exact optimum.
    \item We introduce an efficient caching scheme to accelerate cutting plane
        training.
    \item We demonstrate that using a combination of under-generating and exact
        inference methods, we can learn an SSVM exactly in a practical
        application, even in the presence of loopy graphs.
\end{itemize}

While empirically exact learning yields results comparable to those using
approximate inference alone, certification of optimality allows treating
learning as a black-box, enabling the researcher to focus attention on
designing the model for the application at hand.

\section{Related Work}

Max margin learning for structured prediction was introduced by
\citet{taskar2003max}, in the form of maximum-margin Markov models. Later, this
framework was generalized to structural support vector machines by
\citet{tsochantaridis2006large}. Both works assume tractable loss-augmented
inference.

Currently the most widely used method is the one-slack cutting plane formulation
introduced by \citet{joachims2009cutting}.
This work also introduced the caching of constraints,
which serves as a basis for our work. We improve upon their caching scheme, and in
particular consider how it interacts with approximate inference algorithms.

Recently, there has been an increase in research in learning structured
prediction models where standard exact inference techniques are not applicable,
in particular in the computer vision community.
The influence of approximate inference on structural support vector machine
learning was first analyzed by  \citet{finley2008training}.
\citet{finley2008training} show convergence results for under-generating and
over-generating inference procedures, meaning methods that find suboptimal, but
feasible solutions, and optimal solutions from a larger (unfeasible) set,
respectively.
\citet{finley2008training} demonstrate that over-generating approaches---in
particular linear programming (LP)---perform best on the considered model.
They also show that learning parameters with the LP relaxation minimizes a
bound on the empirical risk when extending the target domain to the relaxed
solutions.
In contrast, we aim at optimizing the non-relaxed objective directly, minimizing
the original empirical risk. This is an important difference, as relaxed solutions
are usually not acceptable in practice.

As using LP relaxations was considered too costly for typical computer vision
approaches, later work employed graph-cuts~\citep{szummer2008learning} or Loopy
Believe Propagation (LBP)~\citep{lucchi2011spatial}. These works use a single
inference algorithm for during the whole learning process, and can not provide any
bounds on the true objective or the empirical risk. In contrast, we combine
different inference methods that are more appropriate for different stages of
learning.

Recently, \citet{meshi2010learning}, \citet{hazan2010primal} and
\citet{komodakis2011efficient} introduced formulations for joint inference and
learning using duality.
In particular, \citet{hazan2010primal} demonstrate the performance of their
model on an image denoising task, where it is possible to learn a large number
of parameters efficiently.
While these approaches show great promise, in particular for pixel-level or
large-scale problems, they perform approximate inference and learning, and do
not relate their results back to the original SSVM objective they approximate.
It is unclear how they compare to standard structured prediction approaches
on real-world applications.

\begin{algorithm*}[t]
    \caption{Cutting Plane Training of Structural SVMs \label{alg_cutting_plane}}

\begin{algorithmic}[1]
    \Require Training samples $\{ (x^{1}, y^{1}), \dots, (x^{n}, y^{n})\}$, regularization parameter $C$, stopping tolerance $\epsilon$, separation oracle $I$.
    \Ensure Parameters $\theta$, slack $\xi$
    \State $\W \leftarrow \emptyset$
    \Repeat
    \State $(\theta, \xi) \leftarrow \displaystyle \argmin_{\theta, \xi} \frac{||\theta||}{2}^2 + C \xi$\label{restricted}\\
    \begin{equation*}
        \text{s.t. }\forall \hat{\mathbf{y}}=(\hat{y}^1, \dots, \hat{y}^n) \in \W:
        \left \langle \theta, \sum_{i=1}^n [\psi(x^i, y^i) - \psi(x^i,
            \hat{y}^i)] \right \rangle \geq \sum_{i=1}^n \Delta(y^i, \hat{y}^i)
            - \xi
    \end{equation*}
    \For {i=1, \dots, n}
    \State
    $\hat{y}^i \leftarrow I(x^i, y^i, \theta) \approx \displaystyle \argmax_{\hat{y}\in\mathcal{Y}} \sum_{i=1}^n \Delta(y^i, \hat{y}) - \left \langle \theta, \sum_{i=1}^n [\psi(x^i, y^i) - \psi(x^i, \hat{y})] \right \rangle$ \label{get_cutting_plane}
    \EndFor
    \State $\W \leftarrow \W \cup \{ (\hat{y}^1, \dots, \hat{y}^n) \} $
    \State $ \displaystyle \xi' \leftarrow  \sum_{i=1}^n \Delta(y^i, \hat{y}^i) - \left \langle \theta, \sum_{i=1}^n [\psi(x^i, y^i) - \psi(x^i, \hat{y}^i)] \right \rangle$
    \Until $\xi' - \xi < \epsilon$ \label{convergence_check}
\end{algorithmic}
\end{algorithm*}

\section{Efficient Cutting Plane Training of SSVMs}\label{learning}

\subsection{The Cutting Plane Method}\label{cutting_plane}

When learning for structured prediction in the max-margin framework of
\citet{tsochantaridis2006large}, predictions are made as
\[ \argmax_{y \in \mathcal{Y}} f(x, y, \theta),\]
where $x \in \mathcal{X}$ is the input, $y \in \mathcal{Y}$ the prediction, and $\theta$ are the parameters to be learned.
We will assume $y$ to be multivariate, $y=(y_1, \dots, y_k)$ with possibly varying $k$.

The function $f$ measures compatibility of $x$ and $y$ and is a linear function of the parameters $\theta$:
\[f(x, y, \theta) = \langle \theta, \psi(x, y) \rangle.\]
Here $\psi(x, y)$ is a joint feature vector of $x$ and $y$. Specifying a particular SSVM model
amounts to specifying $\psi$.

For a given loss $\Delta$, the parameters $\theta$ are learned by minimizing
the loss-based soft-margin objective
\begin{equation}\label{mainequation}
\min_\theta \frac{1}{2} ||\theta||^2 + C \sum_i  r_i(\theta)
\end{equation}
with regularization parameter $C$, where $r_i$ is a hinge-loss-like upper
bound on the empirical $\Delta$-risk:
\[r_i(x^i, y^i, y) = \left [\max_{y \in \mathcal{Y}} \Delta(y^i, y) + \left<\theta, \psi(x^i, y) - \psi(x^i, y^i) \right > \right]_+ \]

We solve the following reformulation of Equation~\ref{mainequation}, known as one-slack QP formulation:
\begin{align}\label{oneslack}
    \min_{\theta, \xi}\ &\frac{1}{2} ||\theta||^2 + C \xi\\
    \text{s.t. }&\forall \hat{\mathbf{y}}=(\hat{y}^1, \dots, \hat{y}^n) \in \mathcal{Y}^n:\\
        &\left \langle \theta, \sum_{i=1}^n [\psi(x^i, y^i) - \psi(x^i,
            \hat{y}^i)] \right \rangle \geq \sum_{i=1}^n \Delta(y^i, \hat{y}^i)
            - \xi
\end{align}
using the cutting plane method described in
Algorithm~\ref{alg_cutting_plane}~\citep{joachims2009cutting}.

The cutting plane method alternates between solving \Eqref{oneslack}
with a working set $\W$ of constraints, and expanding the working set using the
current $\theta$ by finding $\mathbf{y}$ corresponding to the most violated constraint,
using a separation oracle $I$.
We investigate the construction of $\W$ and the influence of $I$
on learning.


Intuitively, the one-slack formulation corresponds to joining all training
samples into a single training example $(\mathbf{x}, \mathbf{y})$ that has no
interactions between variables corresponding to different data points.
Consequently, only a single constraint is added in each iteration of
Algorithm~\ref{alg_cutting_plane}, leading to very small $\W$. We further use
pruning of unused constraints, as suggested by \citet{joachims2009cutting},
resulting in the size of $\W$ being in the order of hundreds for all experiments.

We also use another enhancement of the cutting plane algorithm introduced by
\citet{joachims2009cutting}, the caching oracle. For each training example $(x^i, y^i)$,
we maintain a set $C^i$ of the last $r$ results of loss-augmented inference
(line~\ref{get_cutting_plane} in Algorithm~\ref{alg_cutting_plane}).
For generating a new constraint $(\hat{y}^1, \dotsc, \hat{y}^n)$ we find
\[ 
    \hat{y}^i \leftarrow \argmax_{\hat{y}\in C^i} \sum_{i=1}^n \Delta(y^i, \hat{y}) - \left \langle \theta, \sum_{i=1}^n [\psi(x^i, y^i) - \psi(x^i, \hat{y})] \right \rangle
\]
by enumeration of $C^i$ and continue until line~\ref{convergence_check}.
Only if $\xi' - \xi < \epsilon$, i.e.\ the produced constraint is not violated, we
return to line~\ref{get_cutting_plane} and actually invoke the separation
oracle $I$.

In computer vision applications, or more generally in graph labeling problems,
$\psi$ is often given as a factor graph over $y$, typically using only unary and pairwise functions:
\[ \langle \theta, \psi(x, y) \rangle = \sum_{i=0}^k \langle \theta_u,  \psi_u(x, y_k) \rangle + \sum_{(i, j) \in E} \langle \theta_p, \psi_p(x, y_k, y_l) \rangle, \]
where $E$ a set of pairwise relations. In this form, parameters $\theta_u$ and $\theta_p$ for unary and
pairwise terms are shared across all entries and pairs.
The decomposition of $\psi$ into only unary and pairwise interactions allows
for efficient inference schemes, based on message passing or graph cuts.

There are two groups of inference procedures, as identified in
\citet{finley2008training}: under-generating and over-generating approaches.
An under-generating approach satisfies $I(x^i, y^i, \theta) \in
\mathcal{Y}$, but does not guarantee maximality in line~\ref{get_cutting_plane}
of Algorithm~\ref{alg_cutting_plane}. An over-generating approach on the other
hand, will solve the loss-augmented inference in line~\ref{get_cutting_plane}
exactly, but for a larger set $\hat{\mathcal{Y}} \supset \mathcal{Y}$, meaning
that possibly $I(x^i, y^i, \theta) \notin \mathcal{Y}$.  We will elaborate on
the properties of under-generating and over-generating inference procedures in
Section~\ref{bounds}.

\subsection{Bounding the Objective}\seclabel{bounds}
Even using approximate inference procedures, several statements
about the original exact objective in (\Eqref{mainequation}) can be
obtained.

Let $o_{\W}(\theta)$ denote objective \Eqref{oneslack} with
given parameters $\theta$ restricted to a working set $\W$, as computed in
line~\ref{restricted} of Algorithm~\ref{alg_cutting_plane} and  let
\[
    o^I(\theta) = C\xi' + \frac{||\theta||}{2}^2
\]
when using inference algorithm $I$, that is $o^I(\theta)$ is the approximation of the primal
objective given by $I$. To simplify exposure, we drop the dependency on $\theta$.

Depending on the properties of the inference procedure $I$ used, it is easy to see:
\begin{enumerate}
    \item If all constraints $\hat{y}$ in  $\W$ are feasible, that is generated
        by an under-generating or exact inference mechanism, then $o_{\W}$ is
        an lower bound on the true optimum $o(\theta^*)$.

    \item If $I$ is an over-generating or exact algorithm, $o^I$ is an upper
        bound on $o(\theta^*)$.
\end{enumerate}

We can also use these observations to judge the suboptimality of a given
parameter $\theta$, that is see how far the current objective is from the true
optimum.  Learning with any under-generating approach, we can use 1. to
maintain a lower bound on the objective. At any point during learning, in
particular if no more constraints can be found, we can then use 2., to also
find an upper bound.  This way, we can empirically bound the estimation error,
using only approximate inference.  We now describe how we can further use 1. to
both speed up and improve learning.

\section{Efficient Exact Cutting Plane Training of SSVMs}\seclabel{learning}

\subsection{Combining Inference Procedures}\seclabel{combining}
The cutting plane method described in \Secref{cutting_plane} relies only
on some separation oracle $I$ that produces violated constraints when
performing loss-augmented prediction.

Using any under-generating oracle $I$, learning can proceed as long as a
constraint is found that is violated by more than the stopping tolerance
$\epsilon$.  Which constraint is used next has an impact on the speed of
convergence, but not on correctness. Therefore, as long as an under-generating
method does generate constraints, optimization makes progress on the objective.

Instead of choosing a single oracle, we propose to use a succession of
algorithms, moving from fast methods to more exact methods as training
proceeds. This strategy not only accelerates training, it even makes it
possible to train with exact inference methods, which is infeasible otherwise.

In particular, we employ three strategies for producing constraints,
moving from one to the next if no more constraints can be found:
\begin{enumerate*}
    \item Produce a constraint using previous, cached inference results.
    \item Use a fast under-generating algorithm.
    \item Use a strong but slow algorithm that can certify optimality.
\end{enumerate*}


While using more different oracles is certainly possible, we found
that using just these three methods performed very well in practice.  This
combination allows us to make fast progress initially and guarantee optimality
in the end.
Notably, it is not necessary for an algorithm used as 3) to always produce
exact results. For guaranteeing optimality of the model, it is sufficient that
we obtain a certificate of optimality when learning stops.

\subsection{Dynamic Constraint Selection}\seclabel{caching}
Combining inference algorithm as described in \Secref{combining}
accelerates calls to the separation oracle by using faster, less accurate
methods. On the down-side, this can lead to the inclusion of many constraints
that make little progress in the overall optimization, resulting in much more
iterations of the cutting plane algorithm. We found this particularly problematic
with constraints produced by the cached oracle.

We can overcome this problem by defining a more elaborate schedule to switch
between oracles, instead of switching only if no violated constraint can be
found any more. Our proposed schedule is based on the intuition that we only
trust a separation oracle as long as the current primal objective did not move
far from the primal objective as computed with the stronger inference
procedure.


In the following, we use the notation of \Secref{bounds} and indicate
the choices of oracle $I$ with $Q$ for a chosen inference algorithm and $C$ for
using cached constraints.

To determine whether to produce inference results from the cache or to run the inference algorithm,
we solve the QP once with a constraint from the cache. If the resulting $o^C$ verifies
\begin{equation}\eqlabel{cache_test}
    o^C - o^Q < \frac{1}{2} (o^Q - o_{\W})
\end{equation}
we continue using the caching oracle. Otherwise we run the inference algorithm again.
For testing \Eqref{cache_test}, the last known value of $o^Q$ is used, as recomputing it would defy
the purpose of the cache.

It is easy to see that our heuristic runs inference only $O(\log(o^Q -
o_{\W}))$ times more often than the strategy from \citet{joachims2009cutting} in the
worst case.

\section{Experiments}

\subsection{Inference Algorithms}

As a fast under-generating inference algorithm, we used $\alpha$-expansion
moves based on non-submodular graph-cuts using Quadratid Pseudo-Boolean
Optimization (QPBO)~\citep{rother2007optimizing}.  This move-making strategy
can be seen as a simple instantiation of the more general framework of fusion
moves, as introduced by \citet{lempitsky2010fusion}

For inference with optimality certificate, we use the recently developed
Alternating Direction Dual Decomposition (AD$^3$) method of
\citet{martins2011augmented}. AD$^3$ produces a solution to the linear
programming relaxation, which we use as the basis for branch-and-bound.

\begin{table}
    \begin{center}
    \begin{tabular}{lcc}
        \toprule
                    & Average & Global \\
        \cmidrule{1-3}
    Unary terms only & 77.7& 83.2 \\
    Pairwise model (move making)& 79.6&84.6\\
    Pairwise model (exact)& 79.0 & 84.3\\
        \cmidrule{1-3}
    \citet{ladicky2009associative} & 75.8& 85.0\\
    \citet{gonfaus2010harmony} & 77&  75\\
    \citet{lucchi2013learning} & 78.9& 83.7\\
    \bottomrule
    \end{tabular}
    \end{center}
    \caption{Accuracies on the MSRC-21 Dataset.  We compare a baseline model,
    our exact and approximately learned models and state-of-the-art
    approaches.\tablabel{msrcacc}}
    
\end{table}



\subsection{Image Segmentation}
We choose superpixel-based semantic image segmentation as sample application
for this work.  CRF based models have a history of success in this application,
with much current work investigating models and
learning~\citep{gonfaus2010harmony, lucchi2013learning, ladicky2009associative,
kohli2009robust, krahenbuhl2012efficient}.  We use the MSRC-21 and Pascal VOC 20120 datasets, two
widely used benchmark in the field.

We use the same model and pairwise features for the two datasets.
Each image is represented as a neighborhood graph of superpixels.
For each image, we extract approximately 100 superpixels using 
the SLIC algorithm~\citep{achanta2012slic}.

We introduce pairwise interactions between neighboring superpixels, as is
standard in the literature. Pairwise potentials are founded on two
image-based features: color contrast between superpixels, and relative location
(coded as angle), in addition to a bias term.

We set the stopping criterion $\epsilon=10^{-4}$, though using only
the under-generating method, training always stopped prematurely as no violated
constraints could be found any more.

\subsection{Caching}
First, we compare our caching scheme, as described in \Secref{combining}, with the
scheme of \citet{joachims2009cutting}, which produces constrains from the cache
as long as possible, and with not using caching of constraints at all.  For this experiment,
we only use the under-generating move-making inference on the MSRC-21 dataset. Times until convergence
are 397s for our heuristic, 1453s for the heuristic of
\citet{joachims2009cutting}, and 2661s for using no cache, with all strategies
reaching essentially the same objective.

\Figref{caching} shows a visual comparison that highlights the differences
between the methods. Note that neither $o^Q$ nor $o^C$ provide valid upper bounds on the objective,
which is particularly visible for $o^C$ using the method of \cite{joachims2009cutting}.
Using no cache leads to a relatively smooth, but slow convergence, as inference is run often.
Using the method of \citet{joachims2009cutting}, each run of the separation oracle is followed by
quick progress of the dual objective $o_\W$, which flattens out quickly. Much time is then spent adding
constraints that do not improve the dual solution.
Our heuristic instead probes the cache, and only proceeds using cached constraints if the resulting
$o^C$ is not to far from $o^Q$.

\begin{figure*}
\centering
\includegraphics[width=.7\linewidth]{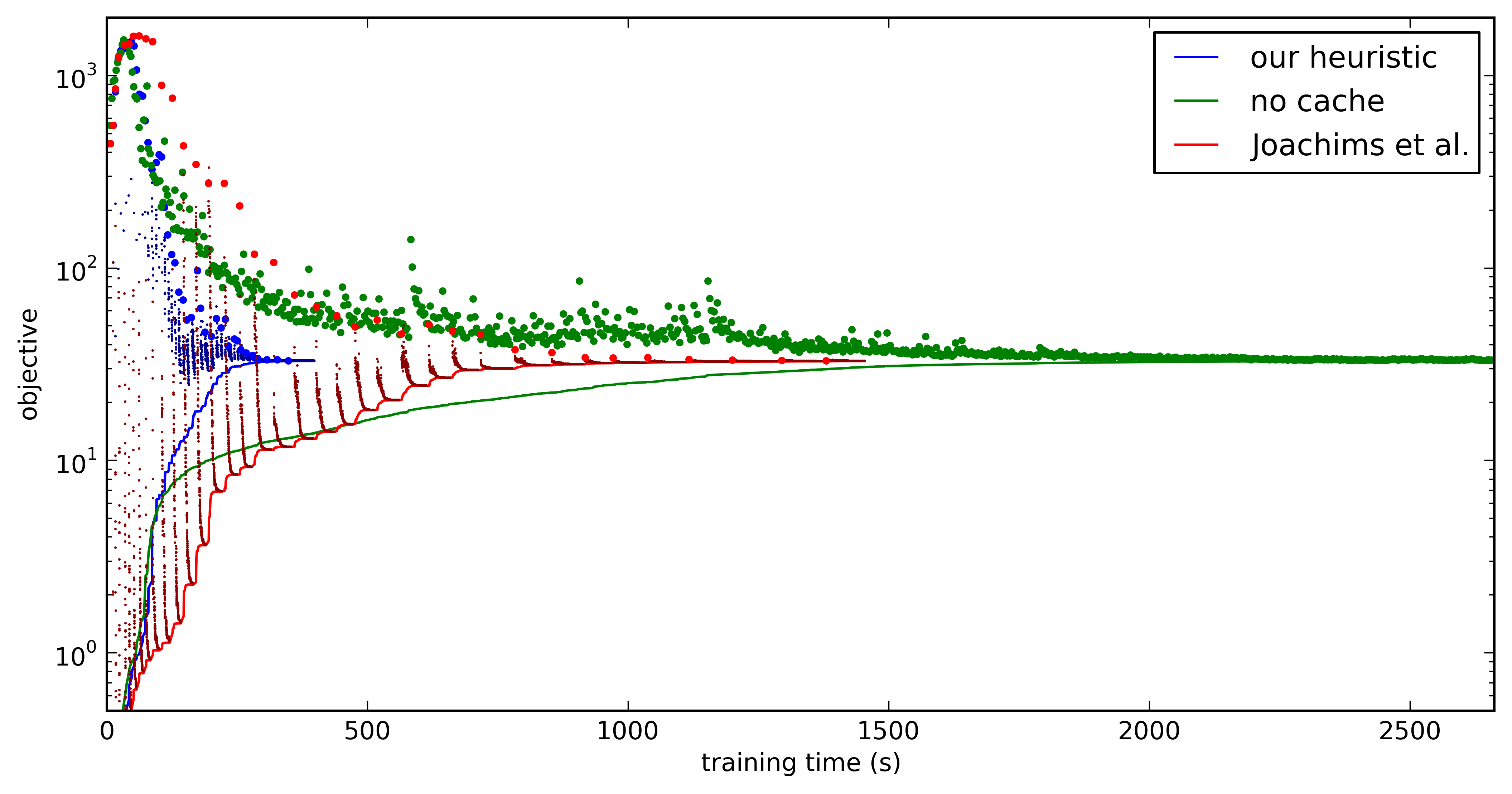}
\caption{%
Training time comparison using different caching heuristics.
Large dots correspond to $o^Q$, small dots correspond to $o^C$,
and the line shows $o_\W$. See the text for details.\figlabel{caching}}
\end{figure*}

\subsubsection{MSRC-21 Dataset}
For the MSRC-21 Dataset, we use unary potentials based on bag-of-words of SIFT
features and color features.  Following \citet{lucchi2011spatial} and
\citet{fulkerson2009class}, we augment the description of each superpixel by a
bag-of-word descriptor of the whole image. To obtain the unary potentials for
our CRF model, we train a linear SVM using the additive $\chi^2$ transform
introduced by \citet{vedaldi2010efficient}. Additionally, we use the unary
potentials provided by \citet{krahenbuhl2012efficient}, which are based on
TextonBoost~\citep{shotton2006textonboost}. This leads to $42 = 2 \cdot 21$
unary features for each node.

The resulting model has around 100 output variables per image, each taking one of 21
labels. The model is trained on 335 images from the standard training and
validation split.

\begin{table}
    \begin{center}
    \begin{tabular}{lc}
        \toprule
                    & Jaccard \\
        \cmidrule{1-2}
    Unary terms only &  27.5 \\
    Pairwise model (move making)& 30.2\\
    Pairwise model (exact) & 30.4\\
        \cmidrule{1-2}
    \citet{dann2012pottics} & 27.4\\
    \citet{krahenbuhl2012efficient} & 30.2\\
    \citet{krahenbuhlparameter} & 30.8\\
    \bottomrule
    \end{tabular}
    \end{center}
    \caption{Accuracies on the Pascal VOC Dataset. We compare our approach
    against approaches using the same unary potentials.\tablabel{pascalacc}}
    
\end{table}

\subsubsection{Pascal VOC 2010}
For the Pascal VOC 2010 dataset, we follow the procedure from \citet{krahenbuhl2012efficient}
in using the official ``validation'' set as our evaluation set, and splitting the training set again.
We use the unary potentials provided by the same work, and compare only against methods
using the same setup and potentials, \citet{krahenbuhlparameter} and \citet{dann2012pottics}.
Note that state-of-the-art approaches, some not build on the CRF framework, obtain
around a Jaccard Index (also call VOC score) of 40\% , notably \cite{xia2012segmentation}, who 
evaluate on the Pascal VOC 2010 ``test'' set.

\begin{table}
    \begin{center}
    \begin{tabular}{lcc}
    \toprule
                    & Move-making & Exact \\
    \cmidrule{1-3}
    Dual Objective $o_\W$ & 65.10 & 67.66  \\
    Estimated Objective $o^I$ &  67.62& 67.66\\
    True Primal Objective $o^E$& 69.92& 67.66\\
    \bottomrule
    \end{tabular}
    \end{center}
    \caption{Objective function values on the MSRC-21 Dataset}
    \tablabel{msrc_objective}
\end{table}

\subsubsection{Results}
We compare classification results using different inference schemes with
results from the literature. As a sanity check, we also provide results without
pairwise interactions.

Results on the MSRC-21 dataset are shown in \Tabref{msrcacc}.
We find that our model is on par with state-of-the-art approaches, implying
that our model is realistic for this task. In particular, our results are comparable to those of
\citet{lucchi2013learning}, who use a stochastic subgradient method with working sets.
Their best model takes 583s for training, while training our model exactly takes 1814s.
We find it remarkable that it is possible to guarantee optimality in time of
the same order of magnitude that a stochastic subgradient procedure with
approximate inference takes. Using exact learning and inference does not increase accuracy
on this dataset.
Learning the structured prediction model using move-making inference alone
takes 4 minutes, while guaranteeing optimality up to  $\epsilon=10^{-4}$
takes only 18 minutes.

Results on the Pascal-VOC dataset are shown in \Tabref{pascalacc}.
We compare against several approaches using the same unary potentials.
For completeness, we also list state-of-the-art approaches not based on CRF models.
Notably, out model matches or exceeds the performance of the much more involved approaches of
\citet{krahenbuhl2012efficient} and \citet{dann2012pottics} which use the same
unary potentials.
Using exact learning and inference slightly increased performance on this dataset.
Learning took 25 minutes using move-making alone and 100 minutes to guarantee optimality
up to $\epsilon=10^{-4}$.
A visual comparison of selected cases is shown in \Figref{visual}.

The objective function values using only the under-generating move-making and
the exact inference are detailed in \Tabref{msrc_objective} and \Tabref{pascal_objective}.
We see that a significant gap between the cutting plane objective and the primal objective
remains when using only under-generating inference.
Additionally, the estimated primal objective $o^I$ using under-generating inference is
too optimistic, as can be expected. This underlines the fact that
under-generating approaches can not be used to upper-bound the primal
objective or compute meaningful duality gaps.

\begin{table}
    \begin{center}
    \begin{tabular}{lcc}
    \toprule
                    & Move-making & Exact \\
    \cmidrule{1-3}
    Dual Objective $o_\W$ &92.06& 92.24\\
    Estimated Objective $o^I$ & 92.07 &92.24\\
    True Primal Objective $o^E$&92.35& 92.24  \\
    \bottomrule
    \end{tabular}
    \end{center}
    \caption{Objective function values on the Pascal Dataset.}
    \tablabel{pascal_objective}
\end{table}

\subsection{Implementation Details}
We implemented the cutting plane Algorithm~\ref{alg_cutting_plane} and our pairwise
model in Python.
Our cutting plane solver for \Eqref{mainequation} uses
\texttt{cvxopt}~\citep{dahl2006cvxopt} for solving the QP in the inner loop. Code for
our structured prediction framework will be released under an open source
license upon acceptance.

We used the SLIC implementation provided by \citet{achanta2012slic} to extract superpixels and
the SIFT implementation in the \texttt{vlfeat} package~\citep{vedaldi08vlfeat}.
For clustering visual words, piecewise training of unary potentials and the
approximation to the $\chi^2$-kernel, we made use of the \texttt{scikit-learn}
machine learning package~\citep{pedregosa2011scikit}.
The move-making algorithm using QPBO is implemented with the help of the QPBO-I
method provided by \citet{rother2007optimizing}.
We use the excellent implementation of AD$^3$ provided by the authors of
\citet{martins2011augmented}. 

Thanks to using these high-quality implementations, running the whole pipeline
for the pairwise model takes less than an hour on a 12 core CPU\@. Solving the
QP is done in a single thread, while inference is parallelized over all cores.
 
\section{Discussion}
In this work we demonstrated that it is possible to learn state-of-the-art CRF models
exactly using structural support vector machines, despite the model containing many loops.
The key to efficient learning is the combination of different inference mechanisms and
a novel caching scheme for the one-slack cutting plane method, in combination
with state-of-the-art inference methods.

We show that guaranteeing exact results is feasible in a practical setting, and
hope that this result provides a new perspective onto learning loopy models for
computer vision applications.
Even though exact learning does not necessarily lead to a large improvement in
accuracy, it frees the practitioner from worrying about optimization and
approximation issues, leaving more room for improving the model, instead of the
optimization.

We do not expect learning of pixel-level models, which typically have tens or
hundreds of thousands of variables, to be efficient using exact inference. However we
believe our results will carry over to other super-pixel based approaches.
Using other over-generating techniques, such as duality-based message passing
algorithms, it might be possible to obtain meaningful bounds on the true
objective, even in the pixel-level domain.

\begin{figure*}
\centering
\includegraphics[width=\linewidth]{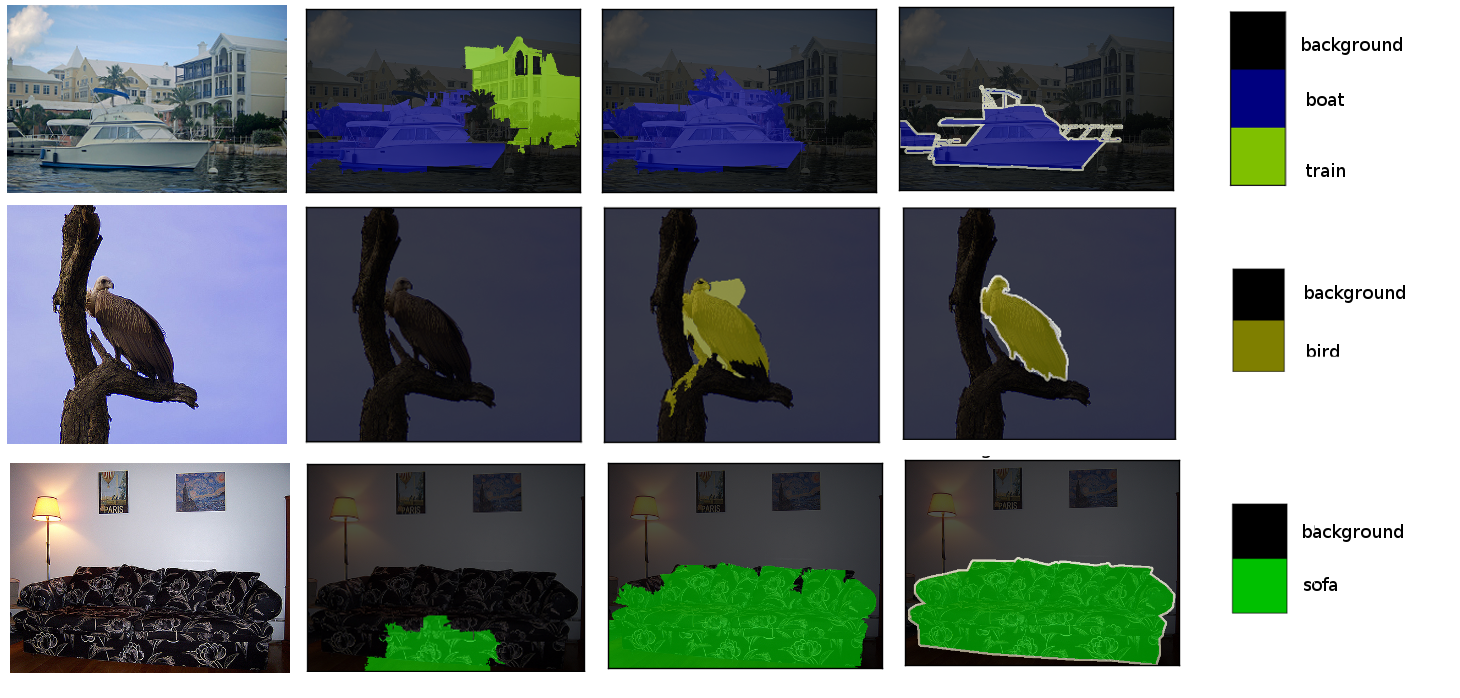}
\caption{%
    Visual comparison of the result of exact and approximate learning on
    selected images from the test set.  From left to right: the input image,
    prediction using approximate learning, prediction using exact learning, and
    ground truth.
\figlabel{visual}}
\end{figure*}
\bibliographystyle{apalike}
\bibliography{paper}

\begin{thebibliography}{}

\bibitem[Achanta et~al., 2012]{achanta2012slic}
Achanta, R., Shaji, A., Smith, K., Lucchi, A., Fua, P., and S{\"u}sstrunk, S.
  (2012).
\newblock {SLIC Superpixels Compared to State-of-the-Art Superpixel Methods}.
\newblock {\em Pattern Analysis and Machine Intelligence}.

\bibitem[Dahl and Vandenberghe, 2006]{dahl2006cvxopt}
Dahl, J. and Vandenberghe, L. (2006).
\newblock Cvxopt: A python package for convex optimization.

\bibitem[Dann et~al., 2012]{dann2012pottics}
Dann, C., Gehler, P., Roth, S., and Nowozin, S. (2012).
\newblock Pottics--the potts topic model for semantic image segmentation.
\newblock In {\em German Conference on Pattern Recognition (DAGM)}.

\bibitem[Finley and Joachims, 2008]{finley2008training}
Finley, T. and Joachims, T. (2008).
\newblock {Training structural SVMs when exact inference is intractable}.
\newblock In {\em International Conference on Machine Learning}.

\bibitem[Fulkerson et~al., 2009]{fulkerson2009class}
Fulkerson, B., Vedaldi, A., and Soatto, S. (2009).
\newblock Class segmentation and object localization with superpixel
  neighborhoods.
\newblock In {\em International Converence on Computer Vision}.

\bibitem[Gonfaus et~al., 2010]{gonfaus2010harmony}
Gonfaus, J., Boix, X., van~de Weijer, J., Bagdanov, A., Serrat, J., and
  Gonzalez, J. (2010).
\newblock Harmony potentials for joint classification and segmentation.
\newblock In {\em Computer Vision and Pattern Recognition}.

\bibitem[Hazan and Urtasun, 2010]{hazan2010primal}
Hazan, T. and Urtasun, R. (2010).
\newblock A primal-dual message-passing algorithm for approximated large scale
  structured prediction.
\newblock In {\em Neural Information Processing Systems}.

\bibitem[Joachims et~al., 2009]{joachims2009cutting}
Joachims, T., Finley, T., and Yu, C.-N.~J. (2009).
\newblock {Cutting-plane training of structural SVMs}.
\newblock {\em Machine Learning}, 77(1).

\bibitem[Kohli et~al., 2009]{kohli2009robust}
Kohli, P., Torr, P.~H., et~al. (2009).
\newblock Robust higher order potentials for enforcing label consistency.
\newblock {\em International Journal of Computer Vision}, 82(3).

\bibitem[Komodakis, 2011]{komodakis2011efficient}
Komodakis, N. (2011).
\newblock Efficient training for pairwise or higher order crfs via dual
  decomposition.
\newblock In {\em Computer Vision and Pattern Recognition}.

\bibitem[Kr{\"a}henb{\"u}hl and Koltun, 2012]{krahenbuhl2012efficient}
Kr{\"a}henb{\"u}hl, P. and Koltun, V. (2012).
\newblock {Efficient inference in fully connected CRFs with Gaussian edge
  potentials}.

\bibitem[Kr{\"a}henb{\"u}hl and Koltun, 2013]{krahenbuhlparameter}
Kr{\"a}henb{\"u}hl, P. and Koltun, V. (2013).
\newblock Parameter learning and convergent inference for dense random fields.
\newblock In {\em International Conference on Machine Learning}.

\bibitem[Ladicky et~al., 2009]{ladicky2009associative}
Ladicky, L., Russell, C., Kohli, P., and Torr, P. (2009).
\newblock {Associative hierarchical CRFs for object class image segmentation}.
\newblock In {\em International Converence on Computer Vision}.

\bibitem[Lempitsky et~al., 2010]{lempitsky2010fusion}
Lempitsky, V., Rother, C., Roth, S., and Blake, A. (2010).
\newblock Fusion moves for markov random field optimization.
\newblock {\em Pattern Analysis and Machine Intelligence}, 32(8).

\bibitem[Lucchi et~al., 2011]{lucchi2011spatial}
Lucchi, A., Li, Y., Boix, X., Smith, K., and Fua, P. (2011).
\newblock Are spatial and global constraints really necessary for segmentation?
\newblock In {\em International Converence on Computer Vision}.

\bibitem[Lucchi et~al., 2013]{lucchi2013learning}
Lucchi, A., Li, Y., and Fua, P. (2013).
\newblock Learning for structured prediction using approximate subgradient
  descent with working sets.
\newblock In {\em Computer Vision and Pattern Recognition}.

\bibitem[Martins et~al., 2011]{martins2011augmented}
Martins, A.~F., Figueiredo, M.~A., Aguiar, P.~M., Smith, N.~A., and Xing, E.~P.
  (2011).
\newblock An augmented lagrangian approach to constrained map inference.
\newblock In {\em International Conference on Machine Learning}.

\bibitem[Meshi et~al., 2010]{meshi2010learning}
Meshi, O., Sontag, D., Jaakkola, T., and Globerson, A. (2010).
\newblock Learning efficiently with approximate inference via dual losses.
\newblock In {\em International Conference on Machine Learning}.

\bibitem[Pedregosa et~al., 2011]{pedregosa2011scikit}
Pedregosa, F., Varoquaux, G., Gramfort, A., Michel, V., Thirion, B., Grisel,
  O., Blondel, M., Prettenhofer, P., Weiss, R., Dubourg, V., et~al. (2011).
\newblock Scikit-learn: Machine learning in python.
\newblock {\em Journal of Machine Learning Research}, 12.

\bibitem[Rother et~al., 2007]{rother2007optimizing}
Rother, C., Kolmogorov, V., Lempitsky, V., and Szummer, M. (2007).
\newblock {Optimizing binary MRFs via extended roof duality}.
\newblock In {\em Computer Vision and Pattern Recognition}.

\bibitem[Shotton et~al., 2006]{shotton2006textonboost}
Shotton, J., Winn, J., Rother, C., and Criminisi, A. (2006).
\newblock Textonboost: Joint appearance, shape and context modeling for
  multi-class object recognition and segmentation.

\bibitem[Szummer et~al., 2008]{szummer2008learning}
Szummer, M., Kohli, P., and Hoiem, D. (2008).
\newblock {Learning CRFs using graph cuts}.

\bibitem[Taskar et~al., 2003]{taskar2003max}
Taskar, B., Guestrin, C., and Koller, D. (2003).
\newblock Max-margin markov networks.
\newblock {\em Neural Information Processing Systems}.

\bibitem[Tsochantaridis et~al., 2006]{tsochantaridis2006large}
Tsochantaridis, I., Joachims, T., Hofmann, T., Altun, Y., and Singer, Y.
  (2006).
\newblock Large margin methods for structured and interdependent output
  variables.
\newblock {\em Journal of Machine Learning Research}, 6(2).

\bibitem[Vedaldi and Fulkerson, 2008]{vedaldi08vlfeat}
Vedaldi, A. and Fulkerson, B. (2008).
\newblock {VLFeat}: An open and portable library of computer vision algorithms.
\newblock \url{http://www.vlfeat.org/}.

\bibitem[Vedaldi and Zisserman, 2010]{vedaldi2010efficient}
Vedaldi, A. and Zisserman, A. (2010).
\newblock Efficient additive kernels via explicit feature maps.
\newblock In {\em Computer Vision and Pattern Recognition (CVPR)}.

\bibitem[Xia et~al., 2012]{xia2012segmentation}
Xia, W., Song, Z., Feng, J., Cheong, L.-F., and Yan, S. (2012).
\newblock Segmentation over detection by coupled global and local sparse
  representations.

\end{thebibliography}

\vfill
\end{document}